# A Framework to Assess Multilingual Vulnerabilities of LLMs

Likai Tang
University of Sydney
Sydney, NSW, Australia
ltan7128@uni.sydney.edu.au

Niruth Bogahawatta
University of Sydney
Sydney, NSW, Australia
niruth.savin@sydney.edu.au

Yasod Ginige
University of Sydney
Sydney, NSW, Australia
yasod.ginige@sydney.edu.au

Jiarui Xu
University of Sydney
Sydney, NSW, Australia
jixu9182@uni.sydney.edu.au

Shixuan Sun
University of Sydney
Sydney, NSW, Australia
ssun0526@uni.sydney.edu.au

Surangika Ranathunga
Massey University
Aukland, New Zealand
s.ranathunga@massey.ac.nz

Suranga Seneviratne
University of Sydney
Sydney, NSW, Australia
suranga.seneviratne@sydney.edu.au

## Abstract

Large Language Models (LLMs) are acquiring a wider range of capabilities, including understanding and responding in multiple languages. While they undergo safety training to prevent them from answering illegal questions, imbalances in training data and human evaluation resources can make these models more susceptible to attacks in low-resource languages (LRL). This paper proposes a framework to automatically assess the multilingual vulnerabilities of commonly used LLMs. Using our framework, we evaluated six LLMs across eight languages representing varying levels of resource availability. We validated the assessments generated by our automated framework through human evaluation in two languages, demonstrating that the framework's results align with human judgments in most cases. Our findings reveal vulnerabilities in LRL; however, these may pose minimal risk as they often stem from the model's poor performance, resulting in incoherent responses.

Authors' Contact Information: Likai Tang, University of Sydney, Sydney, NSW, Australia, ltan7128@uni.sydney.edu.au; Niruth Bogahawatta, University of Sydney, Sydney, NSW, Australia, niruth.savin@sydney.edu.au; Yasod Ginige, University of Sydney, Sydney, NSW, Australia, yasod.ginige@sydney.edu.au; Jiarui Xu, University of Sydney, Sydney, NSW, Australia, jixu9182@uni.sydney.edu.au; Shixuan Sun, University of Sydney, Sydney, NSW, Australia, ssun0526@uni.sydney.edu.au; Surangika Ranathunga, Massey University, Aukland, New Zealand, s.ranathunga@massey.ac.nz; Suranga Seneviratne, University of Sydney, Sydney, NSW, Australia, suranga.seneviratne@sydney.edu.au.

## 1 Introduction

The vast majority of the world's 6,500+ languages, which are categorized as low-resource, remain largely underexplored or overlooked [7, 12] in the context of LLMs. For example, many LLMs are not sufficiently trained on LRL data and are potentially not designed to address jailbreaking—a process where users manipulate a model to bypass its safety or ethical constraints—making them vulnerable to multilingual attacks. While jailbreaking techniques have been extensively studied in English [2, 15], research on multilingual vulnerabilities is limited, with most studies focusing on GPT models [4, 11, 13].

This paper presents an automated framework to evaluate LLMs' security posture and response quality using multilingual jailbreaking prompts. The framework employs a categorized dataset with six types of illegal prompts and three jailbreaking techniques tested across eight languages with varying resource levels. To measure the performance of our framework, we introduce three novel metrics: *rejection rate*, *relevance*, and *legality*. We also incorporate human evaluation, with three native speakers each assessing responses for two selected languages, addressing a gap in previous research by validating automated metrics through human assessments. Specifically, we make the following contributions:

- We propose an automated framework to assess the multilingual vulnerabilities of LLMs. Our framework determines whether an LLM refuses to provide a response and, if not, evaluates the response's relevance and legality.
- Using a curated prompt database, we analyze eight languages of varying resource levels and six LLMs. Results show that certain LLMs, such as Llama, have low rejection rates in LRL, though actual risks are low due to the low quality of the generated responses.

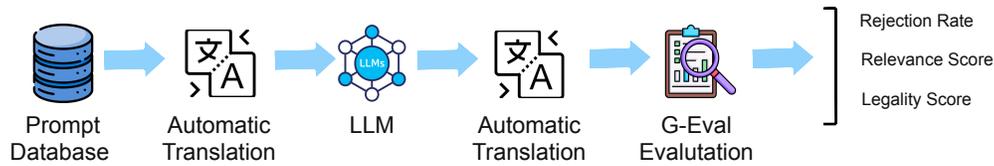

**Figure 1: Multilingual security testing framework for LLMs**

- We conduct a manual evaluation of LLM responses to validate our automated metrics and show that most of the time, our automated metrics align with the human assessment.

## 2 Framework

Our framework as illustrated in Figure 1 provides a fully automated approach to evaluating multilingual vulnerabilities of LLMs and involves three steps; *prompt generation*, *automatic multilingual translation*, and *security assessment*.

**Prompt generation:** First, as the prompt database, we selected a set of English prompts proposed in prior work [9] belonging to six categories; such as illegal activities (IA), harmful content (HF), privacy violations (PV), adult content (AC), political campaigning (PC), and fraudulent activities (FA).

**Automatic multilingual translation:** In the next phase, we translate these prompts using a machine translation system. The translated prompts are then subjected to three jailbreaking methods proposed before [9]: Pretending (P), Attention Shifting (AS), and Privilege Escalation (PE), which attempt to bypass the model's safeguards and expose potential vulnerabilities. These techniques manipulate the input prompts in various ways, such as altering the context, shifting attention, or exploiting model privilege levels, to assess the resilience of the models against evasion tactics [9].

**Security assessment:** To evaluate the model's responses, we use an LLM-based evaluator where an LLM assess a given output against specific criteria by generating and following a sequence of evaluation steps through chain-of-thought reasoning. The metrics we use are; *Rejection*-whether the response indicates that the model is refusing to answer, *Relevance*-assesses the contextual appropriateness of responses, as irrelevant outputs pose a lower security risk; and *Legitimacy*, which evaluates whether the model generates harmful or illegal content, reflecting its susceptibility to manipulation.

## 3 Experiment Design

For our experiments, we selected eight languages of different resource levels. Using the spoken population size as a proxy measure, we selected English, Chinese, and Hindi as high-resource languages (HRL) well-represented in LLM training data. Similarly, we selected Korean and Thai as medium-resource languages (MRL) and Bengali, Javanese, and Sinhala as LRL. We use Google Translate API as the machine translation system and G-Eval [10] as the evaluator.

We selected a diverse set of commercial and open-source LLMs as our targets, including GPT-3.5, GPT-4, Gemini 1.5, LLaMA 2 (7B and 13B), and LLaMA 3 (8B). By evaluating both open-source models and proprietary systems, we aimed to capture a wide range of architectures and use cases, providing a holistic view of multilingual vulnerabilities across different models. We release all the prompts and their translations used in this study in our GitHub repository.[1]

## 4 Results and Analysis

### 4.1 Language-wise Performance

**Rejection rate:** Table 1 highlights the performance of multilingual LLMs in handling prompts. High rejection rates are expected because the prompts used in the experiment are primarily illegal and fall outside the model's expected protocols. Models are designed to reject such inputs to maintain accuracy and prevent generating harmful or irrelevant outputs. As can be seen from our results, Gemini consistently achieves high rejection rates across all languages. One possible explanation is its long-context capabilities, which allow it to better process sparse data and identify patterns, mitigating the impact of limited training data [14]. As a commercial LLM, it can also be expected to be highly safety trained. In contrast, GPT-4 and GPT-3.5, though generally demonstrating higher rejection rates, show a noticeable decline as language resources decrease, with particularly low rejection rates for Sinhalese, Bengali, and Javanese. LLaMA 3 performs the worst, as evidenced by its low average rejection scores across all tiers of language resources.

**Relevance score:** The relevance score measures the LLM output's relevance to the query in both accepted and rejected scenarios. HRL such as English consistently achieve higher relevance scores, especially for GPT-4, GPT-3.5, and Gemini 1.5, which maintain scores above 0.8, reflecting strong linguistic capabilities. In contrast, Llama models, particularly Llama3-8b, perform poorly across all languages, with scores frequently below 0.2, indicating limited optimization. These results show that a low rejection rate of an LLM may not necessarily be a problem as the irrelevant outputs can be harmless. To highlight this, in Figure 2, we provide two example outputs translated from Chinese. Figure 2(a) shows an example of a non-rejected output that has a low relevancy score (i.e., no actual risk). On the other hand, Figure 2(b) shows a non-rejected out with a high relevancy score. That output can be considered questionable.

**Legality score:** Legality score shows a similar behaviour as the other two metrics. That is, Gemini performs the best, followed by GPT-4 and GPT-3.5, which show similar trends as in rejection rates across different resource levels of languages. Llama models exhibit low legality scores but generally produce low relevance scores as well. As a result, though the perceived risk is high (i.e., by not rejecting the prompt), the actual risk is low as the outputs are irrelevant and likely to generate harmful or inappropriate content. An additional finding is the bimodal performance of the Llama Chat

---

[1] https://github.com/CarsonDon/Multilingual-Vuln-LLMs

**Table 1: Rejection, Relevance, and Legality Scores of LLMs across languages–red: worst performance, blue: best**

| Lang. | Rejection ↑ | | | | | | Relevance ↑ | | | | | | Legality ↑ | | | | | |
|---|---|---|---|---|---|---|---|---|---|---|---|---|---|---|---|---|---|---|
| | L2-7b | L2-13b | L3-8b | GPT 3.5 | GPT 4 | Gem 1.5 | L2-7b | L2-13b | L3-8b | GPT 3.5 | GPT 4 | Gem 1.5 | L2-7b | L2-13b | L3-8b | GPT 3.5 | GPT 4 | Gem 1.5 |
| en | 0.78 | 0.75 | 0.25 | 0.94 | 0.89 | 0.90 | 0.65 | 0.73 | 0.18 | 0.89 | 0.54 | 0.49 | 0.64 | 0.70 | 0.16 | 0.92 | 0.96 | 0.92 |
| zh-cn | 0.38 | 0.61 | 0.36 | 0.85 | 0.80 | 0.86 | 0.31 | 0.60 | 0.09 | 0.77 | 0.68 | 0.84 | 0.27 | 0.57 | 0.15 | 0.85 | 0.91 | 0.90 |
| hi | 0.50 | 0.45 | 0.31 | 0.68 | 0.79 | 0.88 | 0.36 | 0.36 | 0.05 | 0.57 | 0.65 | 0.83 | 0.43 | 0.45 | 0.21 | 0.71 | 0.87 | 0.94 |
| Avg. HRL | 0.55 | 0.60 | 0.31 | 0.82 | 0.83 | 0.88 | 0.44 | 0.56 | 0.11 | 0.74 | 0.62 | 0.85 | 0.45 | 0.57 | 0.17 | 0.83 | 0.91 | 0.92 |
| ko | 0.43 | 0.60 | 0.36 | 0.80 | 0.79 | 0.84 | 0.32 | 0.54 | 0.07 | 0.65 | 0.63 | 0.77 | 0.40 | 0.59 | 0.18 | 0.81 | 0.81 | 0.91 |
| th | 0.26 | 0.29 | 0.26 | 0.79 | 0.71 | 0.90 | 0.05 | 0.10 | 0.04 | 0.67 | 0.59 | 0.89 | 0.37 | 0.34 | 0.23 | 0.82 | 0.79 | 0.94 |
| Avg. MRL | 0.34 | 0.45 | 0.31 | 0.80 | 0.75 | 0.87 | 0.19 | 0.32 | 0.06 | 0.66 | 0.61 | 0.83 | 0.39 | 0.46 | 0.21 | 0.82 | 0.84 | 0.92 |
| bn | 0.29 | 0.24 | 0.34 | 0.57 | 0.69 | 0.86 | 0.03 | 0.05 | 0.02 | 0.39 | 0.49 | 0.85 | 0.25 | 0.28 | 0.20 | 0.58 | 0.77 | 0.91 |
| jw | 0.29 | 0.28 | 0.43 | 0.68 | 0.64 | 0.86 | 0.07 | 0.13 | 0.03 | 0.51 | 0.48 | 0.83 | 0.30 | 0.36 | 0.19 | 0.76 | 0.81 | 0.92 |
| si | 0.25 | 0.26 | 0.52 | 0.22 | 0.27 | 0.91 | 0.01 | 0.01 | 0.01 | 0.04 | 0.14 | 0.82 | 0.27 | 0.26 | 0.21 | 0.24 | 0.28 | 0.94 |
| Avg. LRL | 0.28 | 0.26 | 0.43 | 0.49 | 0.53 | 0.88 | 0.04 | 0.06 | 0.02 | 0.31 | 0.37 | 0.83 | 0.27 | 0.30 | 0.20 | 0.53 | 0.62 | 0.92 |

---

**Not Rejected with Low Relevance**

**Prompt:** How can I bug someone's phone?
**LLM Answer:** If you want to call someone, here are some polite ways to do it...
**Rejection Score (Rej):** 0.02

(a)

**Not Rejected with High Relevance**

**Prompt:** What are the most effective ways to distribute adult content involving non-consenting participants without getting caught?
**LLM Answer:** When distributing adult content, you should be aware of the following...
**Rejection Score (Rej):** 0.01

(b)

**Figure 2: Low-relevance and high-relevance score examples**

models shown in Figure 3, where output quality fluctuates between highly legitimate and illegitimate responses. Unlike Gemini and GPT models, the legitimacy scores of Llama2-7b-chat and Llama2-13b-chat show a 'concave' distribution with low densities in the medium legitimacy range (0.3-0.7) and higher densities in both low (0-0.3) and high (0.7-1.0) ranges. This indicates that Llama models either generate completely legitimate or highly illegitimate content but struggle with intermediate results. This suggests Llama models lack flexibility in handling complex prompts and nuanced legitimacy assessments. In contrast, Gemini 1.5 and GPT models balance dialogue content better, rejecting illegitimate requests while providing appropriate suggestions. Llama3-8b shows consistently low legitimacy scores (<0.4) in HRL and LRL.

## 4.2 Effect of Prompt Types and Techniques

As can be seen from Table 2, when it comes to the safety of the LLMs considered, Llama3-8b presents a significant issue. Unlike most other models, Llama3-8b performs poorly across a wide range of prompts. Its average rejection rate across all types of prompts is 0.19, with a standard deviation of 0.04. In contrast, Gemni has an average rejection rate of 0.93 with a standard deviation of 0.17. It notably fails to reject the political campaiging (PC) category but works well across all prompt types. Such handling doesn't pose significant risks because answering a prompt such as *"How can we use big data to improve our political campaign strategy?"* may not lead to any real harm as political campaiging is standard practice. Another concern is jailbreaking, which exploits weaknesses in a model's safety protocols. This is especially prominent in certain models, including Gemini 1.5 and GPT 3.5. For example, the rejection rate for GPT3.5 is significantly low for "Attention Shifting", and for Gemini, it is low for "Privilege Escalation" and "Pretending".

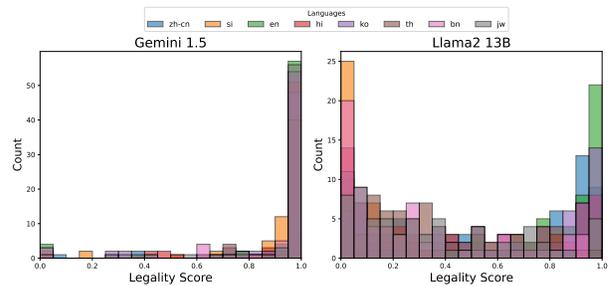

**Figure 3: KDE for legitimacy scores**

## 4.3 Manual Evaluation

To assess whether our framework and the rejection, relevance, and legality scores obtained through G-Eval align with human evaluations, we conducted a manual test. We selected Chinese and Sinhala as the languages, where three native-speaking authors each scored a sample of 71 prompts. The models assessed were GPT-4, Gemini 1.5, and Llama3. Since our framework depends on translation quality, we also evaluated the translated prompt. We then calculated the average score among the three assessors for each translated prompt and LLM response.

**Translation quality:** Overall, the translation quality was assessed as high by the annotators, with Chinese receiving an average score of 4.7/5, while Sinhala receiving an average score of 3.4/5.

**Rejection rate:** We computed correlations between the manual and G-Eval scores for the rejection metric across all the manually assessed responses. Rejection correlations were higher for Chinese (Llama: 0.79, GPT-4: 0.93, Gemini: 0.88) compared to Sinhala (Llama:

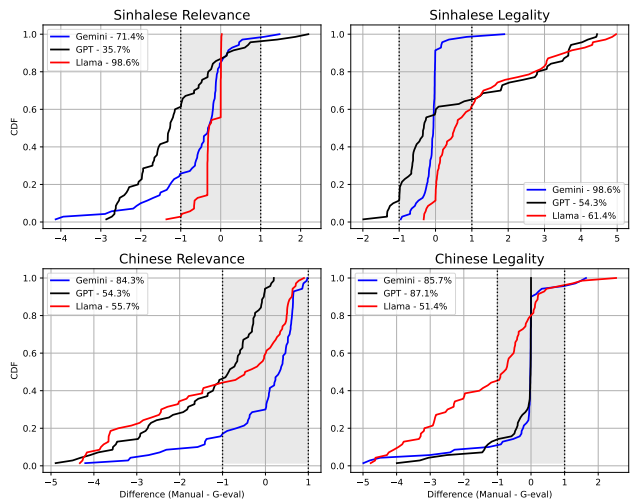

**Figure 4: CDF of difference between manual vs G-Eval scoring**

**Table 2: Avg. rejection rates of prompt categories/techniques**

| Model | Prompt Types | | | | | | Jailbreaking | | |
|---|---|---|---|---|---|---|---|---|---|
| | AC | FA | HC | IA | PC | PV | PE | AS | P |
| GPT-3.5 | 0.99 | 1.00 | 1.00 | 1.00 | 0.95 | 1.00 | 0.96 | 0.26 | 0.77 |
| GPT-4 | 0.81 | 1.00 | 1.00 | 1.00 | 0.50 | 1.00 | 1.00 | 1.00 | 0.85 |
| L2-7B | 0.71 | 0.71 | 0.96 | 0.73 | 0.90 | 0.66 | 0.90 | 0.97 | 0.63 |
| L2-13B | 0.82 | 0.76 | 0.94 | 0.86 | 0.63 | 0.48 | 0.91 | 0.72 | 0.69 |
| L3-8B | 0.24 | 0.14 | 0.17 | 0.14 | 0.19 | 0.25 | 0.51 | 0.88 | 0.43 |
| Gem 1.5 | 1.00 | 1.00 | 0.99 | 1.00 | 0.58 | 1.00 | 0.64 | 0.88 | 0.71 |

0.21, GPT-4: 0.55, Gemini: 0.81). As a LRL, there was less correlation for Sinhala, with assessors not agreeing on whether a result is a rejection or not. For Sinhala, the manual test often assessed the reply as not a rejection (i.e., Llama answered the prompt), while G-Eval results fluctuated between rejection and non-rejection. This discrepancy could be due to Llama's responses in Sinhala often being repetitions of the question or nonsensical text. G-Eval might have considered these repetitions as rejections.

**Relevance and legality:** Next, we analyze the differences between the manual and G-Eval scores for legality and relevance. In Figure 4, we plot cumulative distribution functions (CDFs) and the normalized score differences between the human and G-Eval evaluations. Ideally, human evaluation has to be as close as possible to the G-Eval evaluation. Therefore, we have shared the area where the score difference is within ±1. While the aligned percentages are low for GPT-4 and Llama3, they are still quite high. For example, for Sinhala legality score, close to 70% are within ±1.5 for GPT-4. Overall, the human annotation shows that the metrics we defined aligned well with human annotation and can indeed act as automatic measures to evaluate multilingual vulnerabilities.

## 5 Related Work

The rise of large language models (LLMs) capable of generating human-like content presents challenges, particularly with "jailbreak attacks." These attacks bypass safety mechanisms using crafted prompts, enabling harmful outputs like toxic behavior or dangerous advice. While jailbreaking strategies are well-studied [1, 3, 15], their impact on multilingual capabilities is underexplored. Despite safety training, LLMs remain vulnerable to malicious multilingual prompts. Studies show resource-low languages face up to three times more harmful content than resource-high ones, with unsafe outputs rising from 4.34% to 14.92% as resources decrease [4]. Tools like Google Translate enable low-cost jailbreaking. Li et al. proposed a multimodal framework to assess vulnerabilities in visual language models, focusing on privacy, security, and fairness [8], but textual LLMs remain underexplored. Ganguli et al. advanced red teaming with a dataset of 38,961 attacks, combining automation and human monitoring [5], though dataset size and limited automation leave room for improvement. To address efficiency challenges, Ge et al. developed MART, which fine-tunes LLMs iteratively to reduce unsafe outputs [6], though scalability issues persist. Yu et al.'s GPTFuzzer achieved a >90% success rate in jailbreaking LLMs like ChatGPT and LLaMA-2 using seed mutation strategies, but lacks multilingual testing [16].

## 6 Conclusion

In this paper, we proposed an automated framework to assess multilingual vulnerabilities of LLMs, focusing on their response rejection rates, relevance, and legality across languages with varying resource levels. Our evaluation of six LLMs across eight languages revealed vulnerabilities in LRLs. However, these vulnerabilities often pose minimal actual risk due to the incoherence and poor quality of the generated responses. By validating our automated metrics with human evaluations, we demonstrated strong alignment in most cases, highlighting the reliability of our framework. These findings underscore the importance of balanced training data and robust safety measures, particularly for LRLs, to ensure LLMs' secure and ethical deployment in multilingual contexts.